\documentclass[11pt]{article}
\usepackage[margin=1in]{geometry}
\usepackage[T1]{fontenc}
\usepackage[utf8]{inputenc}
\usepackage{lmodern}
\usepackage{microtype}
\usepackage{booktabs}
\usepackage{tabularx}
\usepackage{enumitem}
\usepackage[hidelinks]{hyperref}
\usepackage{newpxtext,newpxmath}
\usepackage{booktabs, tabularx}
\usepackage[table]{xcolor}

\newcolumntype{L}[1]{>{\raggedright\arraybackslash\hsize=#1\hsize}X}

\title{When Does AI Augment Work?\\
\vspace{8pt}
\large A Workflow-Level Framework for Human-Agent Collaboration}

\author{%
  \parbox{0.98\textwidth}{\centering\normalsize
  Jiaying Wu$^{\dagger}$, Caleb Ziems$^{\dagger}$, \\ Raymond Chan, Nancy F. Chen, Corlyss Chua, Gerard Chung, Jungpil Hahn, Wee Sun Lee, Zhengyuan Liu, Jamie Ng, Desmond C. Ong, Jeryl Ong, Da Ren Soon, Tianqi Song, Tan Zhi-Xuan, Sixing Tao, Emily Yang, Yajing Yang, Stella Xin Yin, Min-Yen Kan$^{\ddagger}$, and Diyi Yang$^{\ddagger}$\thanks{%
    \textit{Author affiliations and contact details.} \\$^{\dagger}$Jiaying Wu and Caleb Ziems contributed as co-first authors, and ${^\ddagger}$Min-Yen Kan and Diyi Yang are the co-organizers of the Collaborative Intelligence and the Future of Work (CIVIC-AI) workshop.\\
    Jiaying Wu: National University of Singapore; \texttt{jiayingwu@u.nus.edu}.\\
    Caleb Ziems: Stanford University; \texttt{cziems@stanford.edu}.\\
    Raymond Chan: Genue; \texttt{raymond@genue.ai}.\\
    Nancy F. Chen: Agency for Science, Technology and Research (A*STAR), Singapore; \texttt{Nancy\_Chen@a-star.edu.sg}.\\
    Corlyss Chua: Ministry of Manpower, Singapore; \texttt{corlyss\_chua@mom.gov.sg}.\\
    Gerard Chung: National University of Singapore, Department of Social Work; \texttt{gerard@nus.edu.sg}.\\
    Jungpil Hahn: National University of Singapore, Department of Information Systems and Analytics; \texttt{jungpil@nus.edu.sg}.\\
    Wee Sun Lee: National University of Singapore, Department of Computer Science; \texttt{dcsleews@nus.edu.sg} \\
    Zhengyuan Liu: Agency for Science, Technology and Research (A*STAR), Singapore; \texttt{liu\_zhengyuan@a-star.edu.sg}.\\
    Jamie Ng: A*STAR Institute of Advanced Intelligence and Computing; \texttt{jamie\_ng@a-star.edu.sg}.\\
    Desmond C. Ong: The University of Texas at Austin, Department of Psychology.\\
    Jeryl Ong: Ministry of Manpower, Singapore; \texttt{jeryl\_ong@mom.gov.sg}.\\
    Tianqi Song: National University of Singapore; \texttt{tianqi\_song@u.nus.edu}.\\
    Da Ren Soon: Ministry of Manpower, Singapore; \texttt{soon\_da\_ren@mom.gov.sg}.\\
    Tan Zhi-Xuan: National University of Singapore, Department of Computer Science; \texttt{xuan.cs@nus.edu.sg}.\\
    Sixing Tao: University of Washington; \texttt{sixint@cs.washington.edu}.\\
    Emily Yang: Independent Researcher; \texttt{hcai@emilyyang.ai}.\\
    Yajing Yang: National University of Singapore, Department of Computer Science; \texttt{yajing.yang@u.nus.edu}.\\
    Stella Xin Yin: Nanyang Technological University, Wee Kim Wee School of Communication and Information; \texttt{xin.yin@ntu.edu.sg}.\\
    Min-Yen Kan: National University of Singapore, Department of Computer Science; \texttt{knmnyn@nus.edu.sg}.\\
    Diyi Yang: Stanford University; \texttt{diyiy@stanford.edu}.\\
  }%
  }
}
\date{}

\date{}
\begin{document}
\maketitle
\begin{abstract}
We aim to characterise the value of artificial intelligence in the workplace. Current studies largely measure this value in terms of the current automation capabilities and public adoption of AI. However, such metrics ignore the greater impacts of human--agent collaboration in transforming the nature of work. To account for this, we must expand the scope of our analysis beyond atomised tasks of today, and instead focus on how AI can augment entire workflows of the future.  To ground this analysis, we establish a precise definition of AI augmentation comprising six conditions, spanning durable net value, meaningful human control, accountability and recovery, and long-term human development through learning, career pathways, and job purpose. We elaborate on these conditions and apply the framework in a case study of AI-mediated social surveys. We conclude by outlining how organisations, researchers, and government leaders can use this framework to make sense of the future of work.
\end{abstract}

\newpage
\section{Introduction}

This whitepaper was written from the discussions and findings of the CIVIC-AI 2026 workshop\footnote{\url{https://civic-ai-collaboration.github.io/2026/}}. Discussions between academia, industry and regulators centred on redesigning work with AI agents and how governance within organisations and by regulators  can support AI agents in augmenting the future of work in the economy. The focus of the whitepaper is to discuss potential recommendations for structuring human--AI interaction at work and the conditions for successful AI augmentation.

\section{Governance Frameworks for Human--AI Collaboration}

Existing AI governance frameworks consistently emphasise human-centred AI, oversight, accountability, and safeguards. For example, the OECD AI Policy Toolkit supports governments in operationalising AI governance principles, while the World Economic Forum (WEF) Human-led AI framework argues that key responsibilities such as direction-setting, judgement, and accountability should remain with humans~\cite{oecd2026toolkit,wef2026asia}. Similarly, earlier guidance from the U.S. Department of Labor \cite{dol2024aibestpractices} and Singapore’s \textit{Model AI Governance Framework}~\cite{imda2026agentic} focus on responsible deployment and risk management.
Existing frameworks increasingly recognise that AI deployment requires redesigning human roles, responsibilities, and governance. However, they provide less guidance on how to evaluate the resulting workflow as a unit of analysis: how task allocation, decision rights, hidden verification work, recovery burdens, and human capability evolve together over time. Moreover, many governance mechanisms, such as disclosure requirements, audit trails, and sign-off protocols, implicitly assume human reviewers retain the capacity to use them, but are silent on whether workflow redesign itself erodes that capacity.  We address both gaps by proposing a \textbf{workflow-level unit of analysis} and by identifying the conditions that distinguish genuine from performative AI augmentation.

\section{A Workflow-Level Analysis of AI-Augmented Work}

Most organisations track the impact of AI in terms of usage logs, investment figures, productivity changes and headcount. These indicators generally measure the scale of AI adoption and its aggregate organisational outcomes. But they do not explicitly measure changes in the nature of work itself --- aggregate outcomes such as headcounts can mask structural changes like team reorganisations, shifting role responsibilities, and the delegation of tasks. Similarly, productivity estimates may omit details on how employees spend their time.  \medskip

We argue that AI will fundamentally change the nature of work through task allocation, hand-offs, and decision rights.  Even with growing recognition that job and workflow redesign is necessary \cite{WorldEconomicForum2026EntryLevelAI}, little work has elaborated how such redesign should be implemented.
We thus argue that workflows are the missing unit of analysis. To measure how AI is restructuring entire workflows, we need to record:

\begin{itemize}
  \item what AI agents may generate, recommend, decide, or execute;
  \item where human workers review, interpret, authorise, override, or intervene; and
  \item how unfamiliar cases, unreliable outputs, and system failures are handled within the workflow.
\end{itemize}

This perspective also makes hidden work visible. While an AI system may reduce the time required to produce an initial output, it may increase the effort required for verification, exception handling or recovery. Such verification costs must be included when accounting for the value of the redesigned workflow. 
We note that structural changes to workflows can also erode human capability, such as when workers' skills atrophy. The integrity of human--AI collaboration drops if humans can no longer reliably oversee AI outputs. That is, human oversight is meaningful only when the reviewer has the competence, time, information, and authority to understand, challenge and override the evolving system. This competence does not persist automatically. It requires organisations to deliberately assign workers enough substantive review work and enough exposure to AI failure to keep the verification skill current. 
In sum, the value of AI in the workplace must be evaluated longitudinally, not merely by measuring task-level productivity, but by examining how human--AI collaboration transforms workflows, roles, and pathways for human development over time.

\begin{table}[h!]
\centering
\small
\renewcommand{\arraystretch}{1.2}
\begin{tabularx}{\textwidth}{@{}>{\raggedright\arraybackslash}p{3.2cm} X >{\raggedright\arraybackslash}p{4.5cm}@{}}
\toprule
\textbf{Condition} & \textbf{Standard} & \textbf{Failure mode} \\
\midrule
\rowcolor{blue!8}
\multicolumn{3}{@{}l@{}}{\textit{Layer 1: Workflow integrity --- snapshot conditions}} \\
\midrule
1. Durable net value~\cite{GonzalezHeidari2025} &
Total multi-stakeholder gain should arise from human--AI complementarity, surviving full accounting of quality, human review, exception handling, rework, recovery, and the cognitive burden shifted to workers. &
Apparent productivity reflects work shifted elsewhere in the workflow. \\
\addlinespace
2. Meaningful human control &
People should retain the competence, time, information, and authority to detect errors, challenge assumptions, constrain agent actions to their intended scope, override decisions, and continue the work during system unavailability. &
Human approval becomes ceremonial, decision fatigue sets in, while deskilling increases the operational risk of unintended outcomes. \\
\addlinespace
3. Clear accountability and recovery &
Decision authority, provenance, escalation, and fallback procedures should be explicitly assigned to human operators. &
Errors lack clear owners, and rare cases or outages disrupt the workflow. \\
\midrule
\rowcolor{orange!12}
\multicolumn{3}{@{}l@{}}{\textit{Layer 2: Human development and job quality --- longitudinal conditions}} \\
\midrule
4. Deepening learning &
Redesign preserves opportunities to develop domain expertise and AI literacy, acquire judgement and responsibility. &
Workers lose important learning opportunities, become dependent on AI outputs. \\
\addlinespace
5. Career pathways &
Job entry points and upward talent progression are preserved or expanded. &
Junior roles are eliminated, job mobility is blocked. \\
\addlinespace
6. Job purpose &
Redesign builds worker motivation and agency, and imbues employment with higher-order meaning. &
AI does desirable tasks, while humans are relegated to undesirable ones. \\
\bottomrule
\end{tabularx}
\caption{Six conditions for genuine AI augmentation.}
\label{tab:augmentation-conditions}
\end{table}

\section{Six Conditions for Genuine Augmentation}

To understand how AI augments human workflows, we need a precise definition of augmentation.  To positively benefit work, AI augmentation should uphold the following six conditions, structured into two logical layers: Layer~1 should be assessed at design, deployment, and periodic operational review. Layer~2 requires longitudinal evidence: repeated observation of how AI-enabled workflows affect workers’ skills, progression, agency, and ability to exercise meaningful control. 
A workflow may satisfy Layer~1 at deployment but fail to remain augmentative if it erodes Layer~2 over time.\medskip

Zooming in on Layer~1's conditions, the appropriate human--agent boundary in a workflow depends on three properties of the task, most closely related to Conditions~2 and 3:

\begin{itemize}
  \item \textbf{Verifiability}: Can a qualified person inspect the output and identify failure?
  \item \textbf{Reversibility}: Can an error or action be corrected before serious harm occurs?
  \item \textbf{Stakes}: What follows if the decision is wrong, delayed, or difficult to contest?
\end{itemize}

Delegation can increase when outputs are easy to inspect, errors are recoverable, and stakes are bounded. Human authority should remain stronger where a task defines goals, interprets ambiguous evidence, establishes validity, or shapes consequential use.
These conditions are interdependent. Output gains lose value when verification consumes the saved time. Human approval offers weak protection when reviewers cannot identify failure. Accountability is incomplete when responsibility is assigned without the information needed to reconstruct a decision. \medskip

Zooming out to Layer~2, we consider longitudinal effects: the interplay between human control, learning and career pathways. The tasks most amenable to AI delegation tend to be high-volume, procedurally defined and verifiable. These are exactly those tasks through which junior workers develop their tacit judgement needed to evaluate outputs, catch failures and push back on algorithmic recommendations. When those tasks are delegated to an AI agent, short-term gains may come at the cost of
developing the human review competence needed over the longer term. A workflow may satisfy all snapshot conditions at deployment yet cease to be genuinely augmentative over time if the human capabilities required for meaningful oversight are not sustained. This is not because oversight procedures have changed but because human workers who will eventually be responsible for oversight lack the formative experiences that make review substantive.\medskip

Meaningful human control is therefore not solely a property of the workflow design but must also be attributed to the larger context of workforce development. It must be actively maintained through task allocation that preserves the learning opportunities that AI would otherwise absorb. Preserving learning and career pathways is not merely an equity concern, but rather a pre-requisite for ensuring human control stays meaningful over time.  



\section{A Worked Example: AI-Mediated Social Surveys}

AI-mediated social surveys show how this boundary can be drawn. Workshop participants placed human researchers in strategic control of the survey purpose, seed questions, weighting, interpretation, and final coding decisions. An AI interviewer such as SparkMe~\cite{anugraha2026sparkme} could tactically revise wording, ask context-sensitive follow-up questions, support data collection, and conduct an initial pass over responses.  This allocation gives the AI agent tasks where scale, responsiveness, and consistency may add value. \medskip

\begin{table}[htbp]
\centering
\small
\renewcommand{\arraystretch}{1.2}
\begin{tabularx}{\textwidth}{@{} L{0.7} L{1.1} L{1.2} @{}}
\toprule
\textbf{Survey stage} & \textbf{Agent role} & \textbf{Human responsibility} \\
\midrule
Purpose and design &
Review wording, identify ambiguities, and flag possible threats to construct validity. &
Define the research purpose, constructs, seed questions, and final instrument. \\
\addlinespace
Data collection &
Conduct adaptive interviews within approved bounds, rephrase questions, and ask relevant follow-ups. &
Approve protocols, monitor sensitive interactions, and inspect question paths. \\
\addlinespace
Processing and coding &
Conduct an initial pass, organise responses, and flag outliers or coding conflicts. &
Resolve disagreements, inspect errors, and approve final coding. \\
\addlinespace
Weighting, analysis, and dissemination &
Support synthesis and exploratory analysis. &
Determine weighting, interpret findings, and authorise consequential use. \\
\bottomrule
\end{tabularx}
\caption{Division of labour in an AI-mediated social survey. At each stage, the agent handles bounded, reviewable execution while humans retain design authority, oversight, and responsibility for consequential decisions.}
\label{tab:survey-division}
\end{table}

Importantly, due to anchoring bias and potential for vacuous verification of an AI’s generation, researchers need to perform the initial designs of protocols and coding themselves.  Such initial coding and protocol design expose junior researchers to opportunities to exercise fine-grained judgement, helping them develop the mastery needed for higher-order strategic decision making.  As workers develop this mastery, they can progress toward higher-order strategic decision making: determining the meaning and validity of the study, what the survey seeks to learn, how concepts are operationalised, how coding conflicts are resolved, and how findings should be interpreted.  \medskip

The boundary is consequential because an AI interviewer becomes part of the measurement instrument. Adaptive questioning may reduce respondent burden, elicit more detailed responses, or reduce some forms of social desirability pressure. It may also introduce hidden assumptions through question reformulation, follow-up selection, coding, or aggregation. Final sign-off provides little assurance when the researcher cannot assess how the evidence was produced. These effects should be measured and assessed as first-class variables. \medskip

A pilot could compare three instruments: a fixed-form survey, a human interviewer, and an adaptive AI interviewer.  A primary snapshot evaluation could include: completion and response rates; respondent burden, response depth, and willingness to disclose; respondent comfort, perceived understanding, privacy concerns, trust, and willingness to participate again; researcher review time; construct validity and internal consistency; and coding accuracy and disagreement. \medskip

Researchers should also retain records of the interview paths produced by the agent. This allows them to inspect how the questions presented to respondents differed from the initial, institutional review board (IRB) approved instrument and whether those changes affected the resulting evidence. Meaningful oversight requires researchers to actively inspect these outputs and paths, revise the instrument, challenge coding decisions, and explain the resulting conclusions. \medskip

A complementary longitudinal evaluation centres on the human researchers, capturing: evolving level of mastery of the fundamental survey coding constructs; junior talent creation and senior talent development; job satisfaction and purpose.
Generalising this case study yields a design rule:

\begin{quote}
    \textit{Assign agents work where gains are scalable, actions are reversible and errors remain inspectable. Retain human authority where decisions set goals, validity, interpretation, or consequential use. Architect the workflow to account for and measure longitudinal effects on human efficacy.}
\end{quote}

\section{Operationalising the Augmentation Framework: \\Singapore as an Application Context}

Singapore's labour market research provides a strong baseline for monitoring AI adoption. The Manpower Research and Statistics Department (MSRD) of the Ministry of Manpower already conducts representative labour-market surveys, integrates survey findings with administrative data, and works with government agencies, academia,
industry, unions, and employers. 
A credible augmentation claim connects headline outcomes with the workflows producing them, addressing each of the six conditions with evidence. Judged by the six conditions, how does Singapore’s current labour market research fare? \medskip

The current evidence supports cautious optimism, but remains incomplete. AI adoption is still at an early stage: 28.5\% of firms reported having adopted AI, and among adopters, 70.7\% reported improvements in worker productivity. Firms more often reported role redesign (18.9\%) than reduced headcount (6.2\%) \cite{mom2026aiadoption}. These observations provide preliminary evidence relevant to Condition~1 (Durable net value), but do not establish it under our definition, as existing statistics do not capture the full costs of verification, exception handling, rework, recovery, or unofficial AI use.

Evidence for the worker-centric conditions is thinner. Entry-level PMET job openings increased slightly from 32,500 in December 2025 to 32,800 in March 2026, while fresh-graduate employment outcomes have remained broadly resilient \cite{mom_fresh_graduate_2026}. These snapshots do not yet indicate broad-based deterioration in entry-level opportunities, but neither do they establish the longer-term effects of AI on human development. Assessing whether AI-enabled workflows deepen learning (Condition~4), preserve career pathways (Condition~5), and sustain agency and job purpose (Condition~6) therefore requires repeated longitudinal measurement.
\medskip

Current measures in Singapore also leave an open question about whether a particular workflow redesign meets our conditions. Also, condition outcomes do not hold equally for all workers. Meaningful human control depends on skills and access that are unevenly distributed: for example, among young workers in Singapore, only 38\% of those with secondary qualifications used new technology at work, against 74\% of degree holders \cite{tan2026technology}. Human control should therefore be assessed as a capability workers can actually exercise, not only as a formal role assigned on paper. 

\section{Moving Forward: Workflow Records} 

A shared workflow record provides a practical basis for collecting evidence to meet auditing requirements for the six conditions. 
Each record should document the baseline process and intended objective; the tasks assigned to the agent and its level of authority; human review, override, and escalation points; exception handling, fallback, and recovery procedures; significant errors and unapproved or unintended agent actions; total effort, including verification and repair; and effects on human capability, AI literacy, learning opportunities, progression, agency, and satisfaction. \medskip

Workflow records should be complemented by employee-level evidence, particularly where formal adoption data may miss unofficial AI use. We emphasise triangulation across employer and employee surveys, administrative records, job-posting data, expert input, industry interviews, and qualitative evidence. These sources capture different aspects of adoption and can reveal effects that aggregate measures obscure. 
A common workflow record can support comparison without imposing the same operating model across sectors.  Sector-specific thresholds still need to apply, as clinical processes, research workflows, public surveys, and client-onboarding systems differ in their requirements for quality, privacy, contestability, and risk.    \medskip

A practical next step is to incorporate workflow modules into organisational pilots and relevant labour-market research. Organisations could use these modules when deciding whether to expand, revise, or discontinue a deployment. Researchers could compare workload, capability, and career-stage effects over time. Public authorities could connect selected workflow measures with employee surveys, graduate outcomes, hiring, and training data.   \medskip

Workflow-level measures complement these existing sources by showing how reported adoption and workforce outcomes arise from changes in the organisation of work. The aim is to make augmentation claims testable: reported gains should include the full cost of human work, oversight should remain effective in practice, failures and unintended actions should be traceable and recoverable, and effects on learning and progression should be measured over time.

\section{Conclusion}

AI changes work through the workflows that incorporate it. Whether such change constitutes genuine augmentation depends on how human and AI capabilities are combined: how tasks and decision authority are allocated, how errors and failures are handled. A workflow-level framework gives organisations a basis for deployment decisions and gives researchers, educators, and public authorities a common object of analysis. Its central proposition is simple:

\begin{quote}
    \textit{AI genuinely augments work when the redesigned workflow creates durable net value, preserves meaningful human control, includes clear accountability and recovery procedures, and sustains viable pathways for learning, progression and purpose.}   
\end{quote}

\section*{Acknowledgements}
This whitepaper synthesises discussions held during Day 2 of the CIVIC-AI 2026 workshop. We thank the workshop participants for contributing examples, questions, and perspectives that informed the analysis. The framing and recommendations are the responsibility of the authors and should not be interpreted as an official position of any participating organisation or institution.


\bibliographystyle{plain}
\bibliography{references}

@misc{mom2026aiadoption,
  author       = {{Manpower Research and Statistics Department, Ministry of Manpower, Singapore}},
  title        = {Adoption of Artificial Intelligence Among Firms},
  year         = {2026},
  publisher    = {Ministry of Manpower},
  url          = {https://stats.mom.gov.sg/Pages/Report-Adoption-of-Artificial-Intelligence-Adoption-Among-Firms.aspx},
  urldate      = {2026-08-25}
}

@article{anugraha2026sparkme,
  title={Sparkme: Adaptive semi-structured interviewing for qualitative insight discovery},
  author={Anugraha, David and Padmakumar, Vishakh and Yang, Diyi},
  journal={arXiv preprint arXiv:2602.21136},
  year={2026}
}

@techreport{WorldEconomicForum2026EntryLevelAI,
  author      = {{World Economic Forum}},
  title       = {Artificial Intelligence and the Future of Entry-Level Work: A Framework for Safeguarding and Reinventing Early Career Pathways},
  institution = {World Economic Forum},
  type        = {Report},
  year        = {2026},
  month       = jun,
  day         = {22},
  url         = {https://www.weforum.org/publications/artificial-intelligence-and-the-future-of-entry-level-work-a-framework-for-safeguarding-and-reinventing-early-career-pathways/}
}

@article{GonzalezHeidari2025,
  author  = {Gonzalez, Cleotilde and Heidari, Homa A.},
  title   = {A Cognitive Approach to Human--{AI} Complementarity in Dynamic Decision-Making},
  journal = {Nature Reviews Psychology},
  volume  = {4},
  pages   = {808--822},
  year    = {2025},
  doi     = {10.1038/s44159-025-00499-x},
  url     = {https://doi.org/10.1038/s44159-025-00499-x}
}

@misc{oecd2026toolkit,
  author       = {{Organisation for Economic Co-operation and Development}},
  title        = {The {OECD} {AI} Policy Toolkit: Better {AI} Policies for Better Lives},
  year         = {2026},
  publisher    = {OECD.AI},
  url          = {https://oecd.ai/en/wonk/the-oecd-ai-policy-toolkit-better-ai-policies-for-better-lives},
  urldate      = {2026-08-25}
}

@misc{wef2026asia,
  author       = {{World Economic Forum}},
  title        = {Asia's Human-led {AI} Opportunity: A Framework for Transformation},
  year         = {2026},
  publisher    = {World Economic Forum},
  url          = {https://www.weforum.org/publications/asia-s-human-led-ai-opportunity-a-framework-for-transformation/},
  urldate      = {2026-08-25}
}

@misc{dol2024aibestpractices,
  author       = {{U.S. Department of Labor}},
  title        = {Department of Labor Releases {AI} Best Practices Roadmap for Developers and Employers, Building on {AI} Principles for Worker Well-Being},
  year         = {2024},
  publisher    = {U.S. Department of Labor},
  url          = {https://www.dol.gov/newsroom/releases/osec/osec20241016},
  urldate      = {2026-08-25}
}

@misc{imda2026agentic,
  author       = {{Infocomm Media Development Authority}},
  title        = {Model {AI} Governance Framework for Agentic {AI}},
  year         = {2026},
  publisher    = {IMDA, Singapore},
  url          = {https://www.imda.gov.sg/resources/press-releases-factsheets-and-speeches/factsheets/2026/updated-model-ai-governance-framework-for-agentic-ai},
  urldate      = {2026-08-25}
}

@techreport{tan2026technology,
  author       = {Tan, Z. H. and Deepa, V. and Ng, I. Y. H. and Yusof, M. Z. B. and Chung, G.},
  title        = {Technological Attitudes, Devices, and Skills at Work and at Home: Report from {In-Work Poverty and the Challenges of Getting By Among the Young}},
  year         = {2026},
  institution  = {Social Service Research Centre, National University of Singapore},
  url          = {https://fass.nus.edu.sg/ssr/wp-content/uploads/sites/8/2026/05/IWP-Tech-Report-202605.pdf},
  urldate      = {2026-08-25}
}

@misc{mom_fresh_graduate_2026,
  author       = {{Ministry of Manpower, Singapore}},
  title        = {Written Answer to {PQ} on Fresh Graduate Employment},
  year         = {2026},
  month        = aug,
  url    = {https://www.mom.gov.sg/newsroom/parliament-questions-and-replies/2026/0804-written-answer-to-pq-on-fresh-graduate-employment},
  urldate      = {2026-08-31}
}

\end{document}